\documentclass[10pt,twocolumn,letterpaper]{article}

\usepackage[pagenumbers]{cvpr} 

\definecolor{cvprblue}{rgb}{0.21,0.49,0.74}
\usepackage[pagebackref,breaklinks,colorlinks,allcolors=cvprblue]{hyperref}
\usepackage{tabularray}
\def\paperID{9968} 
\def\confName{CVPR}
\def\confYear{2025}

\title{CMMLoc: Advancing Text-to-PointCloud Localization with Cauchy-Mixture-Model Based Framework}

\author{Yanlong Xu\textsuperscript{1} \quad Haoxuan Qu\textsuperscript{2} \quad Jun Liu\textsuperscript{2} \quad Wenxiao Zhang\textsuperscript{3}\footnotemark[2] \quad Xun Yang\textsuperscript{1}\footnotemark[2]\\
\textsuperscript{1} University of Science and Technology of China\quad
\textsuperscript{2} Lancaster University\quad
\textsuperscript{3} Hohai University\\
{\tt\small kc30@mail.ustc.edu.cn,\{h.qu5,j.liu81\}@lancaster.ac.uk,wenxxiao.zhang@gmail.com,xyang21@ustc.edu.cn}
}

\begin{document}
\maketitle
\footnotetext[2]{Corresponding authors.}
\begin{abstract}
The goal of point cloud localization based on linguistic description is to identify a 3D position using textual description in large urban environments, which has potential applications in various fields, such as determining the location for vehicle pickup or goods delivery. 
Ideally, for a textual description and its corresponding 3D location, the objects around the 3D location should be fully described in the text description. However, in practical scenarios, e.g., vehicle pickup, passengers usually describe only the part of the most significant and nearby surroundings instead of the entire environment. In response to this \textbf{partially relevant} challenge, we propose \textbf{CMMLoc}, an uncertainty-aware \textbf{C}auchy-\textbf{M}ixture-\textbf{M}odel (\textbf{CMM}) based framework for text-to-point-cloud \textbf{Loc}alization. To model the uncertain semantic relations between text and point cloud, we integrate CMM constraints as a prior during the interaction between the two modalities. We further design a spatial consolidation scheme to enable adaptive aggregation of different 3D objects with varying receptive fields. To achieve precise localization, we propose a cardinal direction integration module alongside a modality pre-alignment strategy, helping capture the spatial relationships among objects and bringing the 3D objects closer to the text modality. Comprehensive experiments validate that CMMLoc outperforms existing methods, achieving state-of-the-art results on the KITTI360Pose dataset. Codes are available in this GitHub repository \url{https://github.com/kevin301342/CMMLoc}.
\vspace{-1.8em}
\end{abstract}

\begin{figure}
    \centering
    \includegraphics[width=0.5\textwidth]{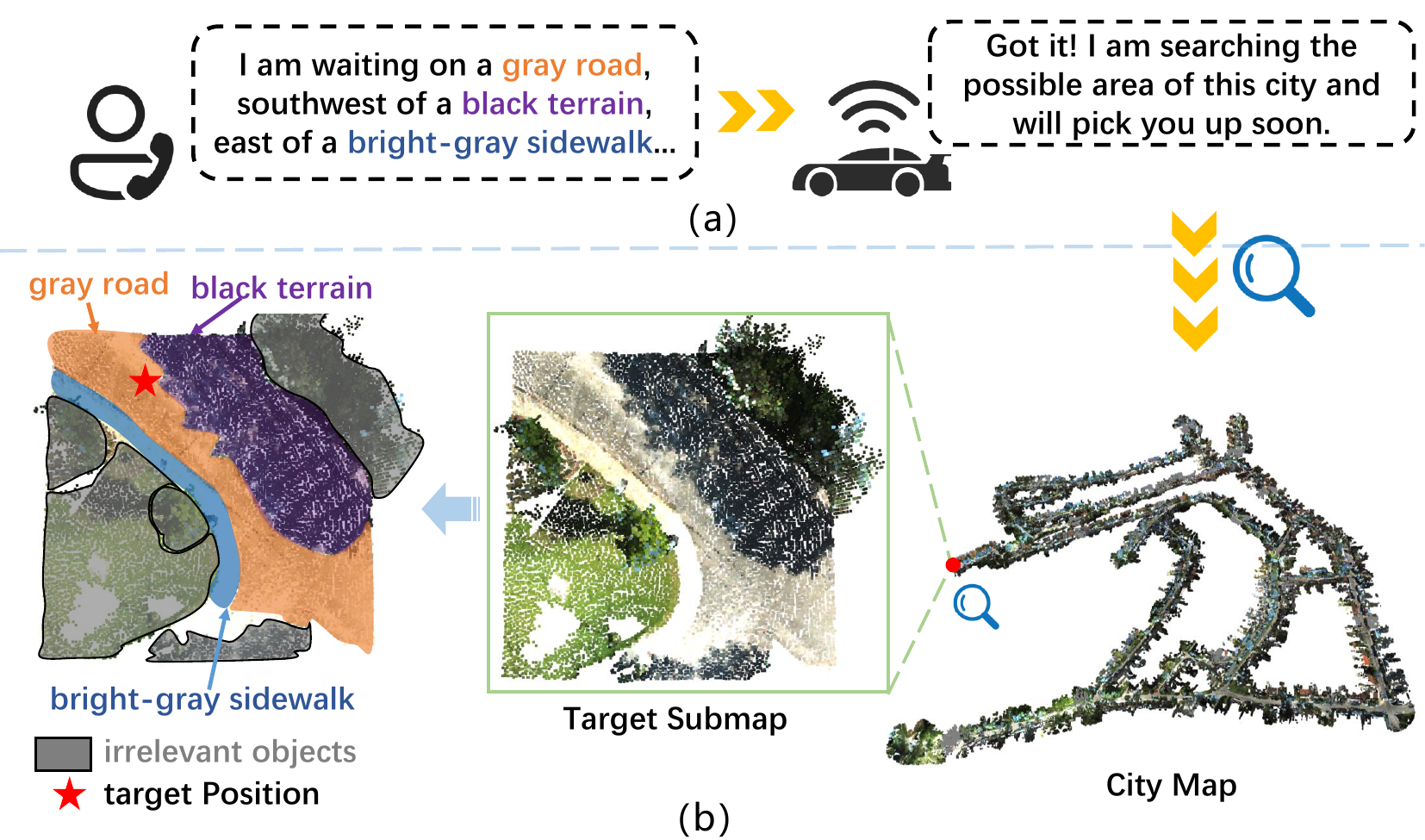}
    \caption{Given a text description of a location in (a), CMMLoc searches the 3D city and identifies the most likely target location of the described position within a submap in (b). Notably, text descriptions do not correspond to all objects within the submap, where the irrelevant objects are shown in gray color.} 
    \label{fig:motivation}
    \vspace{-2em}
\end{figure}

\section{Introduction}
\label{sec:intro}
3D point cloud localization according to the given natural language descriptions, is promising in the future with the development of autonomous driving \cite{hu2023planning} and embodied agents \cite{le2024fast}, especially in large-scale urban environments. In practical scenarios, GPS performance often suffers due to signal blockage in certain environments, such as in urban canyons \cite{vicek1993gps,cui2003autonomous}. Conversely, language-based point cloud localization can help under these conditions. Furthermore, this technology offers significant convenience for humans, as it eliminates the need to provide an exact address, freeing us from the limitations of GPS tags and enabling more user-friendly human-machine interactions.

3D point cloud localization based on text descriptions is a challenging task due to the gap between linguistic descriptions and large-scale point clouds. To simplify this difficult task, the pioneering work Text2Pos \cite{kolmet2022text2pos} proposes a coarse-to-fine pipeline that divides the city map into submaps, first retrieving candidate submaps based on global descriptor matching, followed by fine localization through hint-to-instance correspondence. However, it does not account for the relationships within text queries or between point clouds. RET \cite{wang2023text} improves cross-modal collaboration by introducing a Relation-Enhanced Transformer. Text2Loc \cite{xia2024text2loc} adopts a frozen pre-trained T5 \cite{raffel2020exploring} model and a hierarchical transformer to analyze the relationships between text descriptions with a matching-free regression method for fine localization. 


While these approaches make significant progress on this task, they overlook the \textit{partial relevance characteristic} between text descriptions and the 3D objects in the scene. 
As is shown in \cref{fig:motivation}, during vehicle pickup in (a), the ideal scenario would involve passengers providing a detailed and comprehensive description of all surrounding objects within the target submap in (b), including a gray road, a black terrain, a bright-gray sidewalk, and other irrelevant objects in gray color. However, passengers typically describe only the most significant and nearby surroundings like (a). This selective description introduces uncertainty into the text-to-3D object-matching process, potentially disrupting the semantic modeling between text and 3D objects.  

Based on this observation, we propose CMMLoc, an uncertainty-aware text-to-point-cloud localization framework based on the Cauchy-Mixture-Model. Notably, CMM naturally adapts to partially relevant problems, where both relevant and irrelevant objects are present. This is because, as analyzed in \cite{kalantan2019quantile}, CMM has the property of diminishing the influence of irrelevant objects without entirely disregarding them. This aligns well with the expectations for partially relevant problems.
Our framework employs a coarse-to-fine localization pipeline: initial text-submap retrieval, followed by fine localization. In the coarse submap retrieval stage, we propose a Cauchy-Mixture-Model-based Transformer to model 3D object representations. Inspired by advancements in the NLP field \cite{gmmweight}, we design a Cauchy weighted attention scheme, whose attention weights are attenuated according to the semantic similarity between correlated objects. Specifically, we initially utilize multi-scale Cauchy windows, which incorporate Cauchy distribution as priors, to model the semantic relationships among 3D point cloud objects to generate 3D object features with varying receptive fields. Subsequently, considering the sparsity and various shapes of point clouds, we propose an innovative spatial consolidation scheme that could adaptively aggregate the 3D object features with different receptive fields. In the fine localization stage, we incorporate a cardinal direction integration module combined with a cross-modality pre-alignment strategy. In particular, we first pre-align the multi-modal encoder to bring the 3D objects closer to the text modality. Then we design a Cardinal Direction Integration module to capture spatial relations among objects, which can encourage more fine-grained interactions between 3D objects and text queries.

We conduct extensive experiments on the KITTI360Pose dataset, demonstrating that our framework achieves state-of-the-art performance in cross-modal text-to-point-cloud localization. Additional experimental results further validate the necessity and correctness of modeling the partial relevance between text queries and 3D objects.

In summary, our main contributions are as follows: 
\begin{itemize}
    \item We deeply analyze the current coarse-to-fine pipeline for 3D language-based point cloud localization task and point out the partially relevant characteristic of this task which was overlooked by previous work. To tackle this problem, we propose a novel framework CMMLoc. 
    \item For the coarse stage, we introduce a CMM-based Transformer that incorporates CMM priors for 3D object modeling, along with a spatial consolidation scheme for better submap representation. For fine localization, we propose a pre-alignment strategy and a cardinal direction integration module to enhance cross-modality interactions,  yielding better localization accuracy.   
    \item Extensive experiments on the KITTI360Pose dataset demonstrate the effectiveness of our approach, which achieves state-of-the-art performance.  
\end{itemize}

\begin{figure*}[t]
    \vspace{-2.5em}
    \centering
    \includegraphics[width=1.0\textwidth]{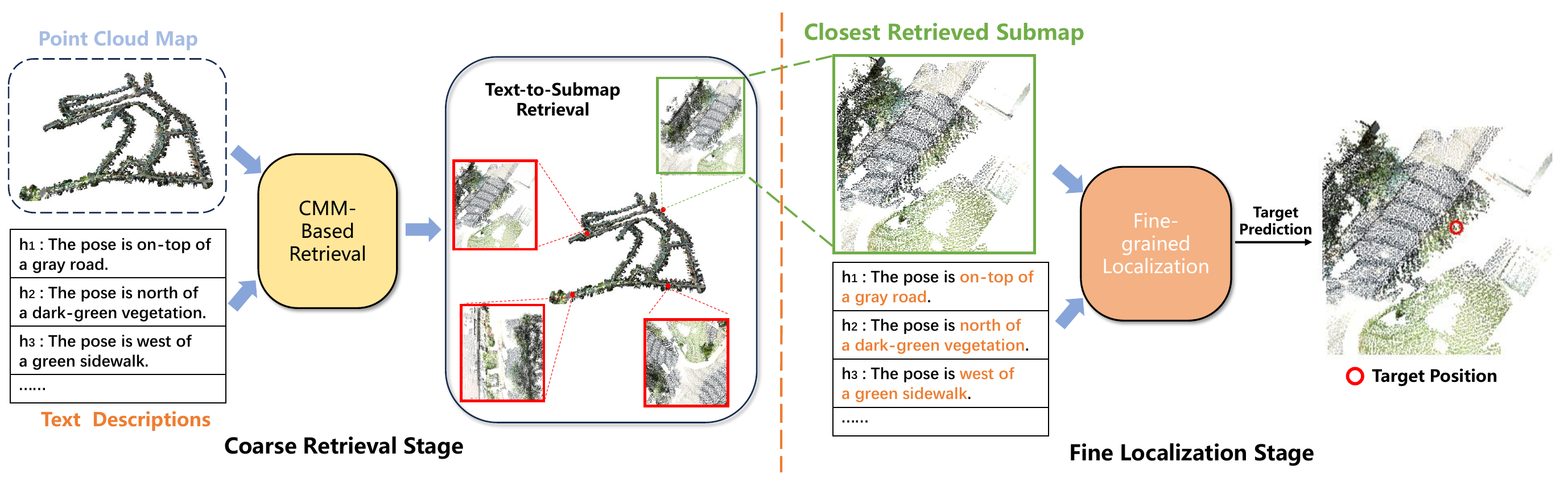}
    \caption{The overview of proposed CMMLoc. It is a coarse-to-fine architecture consisting of two stages: Coarse submap retrieval and Fine localization. \textit{Coarse submap retrieval}. Given text descriptions, we first identify a set of candidate submaps potentially containing the target position. This is achieved by retrieving the Top-k nearest submaps from a constructed database of submaps using our CMM-based retrieval model. \textit{Fine localization}. We then refine the coordinates of the retrieved submaps via our pre-alignment strategy and cardinal direction integration module to improve localization accuracy.}
    \label{fig:overview}
    \vspace{-1em}
\end{figure*}

\section{Related Work}
\label{sec:related}
\textbf{2D visual grounding.} 2D visual grounding aims to estimate an accurate position based on a given image or image query. Traditional techniques cast this problem as an instance retrieval task and employ various aggregation methods to extract invariant image features and then perform objects matching across different viewpoints like Scale Invariant Feature Transform (SIFT) \cite{lowe2004distinctive}, Vector of Locally Aggregated Descriptors (VLAD) \cite{arandjelovic2016netvlad}, Generalized-Mean (GeM) \cite{radford2021learning} and so on. Recent methods \cite{sarlin2019coarse,sattler2016efficient} mostly adopt a coarse-to-fine localization pipeline. Given an image query, the coarse stage uses K-Nearest Neighbors to find candidate subsets, and the fine stage establishes pixel-wise correspondences between the query and candidates for precise position prediction. Compared to 2D visual grounding, our task is more challenging due to the huge gap between text descriptions and large-scale urban point clouds, which is similar with other language-based visual understanding tasks~\cite{yang2021deconfounded,yang2022video,yang2024robust}. The key is to accurately model the cross-modality relationship between different modality instances~\cite{yang2024learning,yang2020tree}.\\
\textbf{3D point cloud based place recognition.} Point cloud-based place recognition is similar to 2D visual grounding, the difference lies in that the query of this task is a single scan from 3D LiDAR sensor rather than an image. Methods for this task can be divided into two categories: those based on global descriptors and those based on plane or object descriptors. Early works like \cite{magnusson2009automatic,rohling2015fast} use global statistics to construct handcrafted global descriptors. Representing the scan as the bird's-eye view (BEV) is another popular approach. Scan Context \cite{kim2018scan}, for example, computes a hand-crafted descriptor from a BEV scan representation. Then deep learning of global descriptors becomes the focus of the research. The first notable method PointNetVLAD \cite{uy2018pointnetvlad} combines PointNet \cite{qi2017pointnet} with NetVLAD \cite{arandjelovic2016netvlad} to aggregate local features into a global descriptor. Following methods like \cite{ zhang2019pcan, sun2020dagc, liu2019lpd,du2020dh3d,xia2021soe} focus on better feature combination.  Transformer-based methods like \cite{barros2022attdlnet,zhou2021ndt,deng2018ppfnet,fan2022svt,ma2022overlaptransformer,ma2023cvtnet} utilize different transformer networks~\cite{pan2023finding} to enhance the performance. However, MinkLoc3D \cite{komorowski2021minkloc3d} employs sparse 3D convolution and outperforms transformer-based methods with fewer parameters. MinkLoc++ \cite{komorowski2021minkloc++} further improves MinkLoc3D by incorporating the ECA \cite{wang2020eca} attention mechanism, similar to Transloc3D \cite{xu2021transloc3d}. Recently, the re-ranking technique has been taken into consideration by researchers like \cite{zhang2022rank} to boost efficiency. There also exist works like \cite{fernandez2013fast, finman2015toward} which propose to use 3D shapes or object-centric recognition as a different solution. In contrast to point cloud place recognition, our task uses linguistic descriptions as queries to locate the target position.\\
\textbf{Text-guided point cloud localization.} Given a 3D point cloud scene, text-guided point cloud localization refers to the task of semantically locating a target position based on language descriptions. It was first introduced in \cite{chen2020scanrefer,achlioptas2020referit3d}. Recent research in this area can be broadly classified into two categories: two-stage and one-stage frameworks. Two-stage frameworks like \cite{roh2022languagerefer,chen2022language,he2021transrefer3d} first generate multiple proposals by the pre-trained 3D detector and then reason about detected proposals with text descriptions to output the final result. In contrast, one-stage methods like \cite{huang2023dense,luo20223d,wu2023eda} take the whole 3D scene as input and directly regress the spatial bounding box of the text-guided object. However, most of these works focus on indoor scenes, whereas our work is concerned with large-scale outdoor scene localization. Text2Pos \cite{kolmet2022text2pos} is the first to formalize this task and introduce the first dataset along with a coarse-to-fine localization pipeline. RET \cite{wang2023text} and Text2Loc \cite{xia2024text2loc} continue to improve the performance with their novel designs. In this work, we also follow the two-stage pipeline but emphasize the partial relevance characteristic of this task, which has been overlooked in previous studies. 



\section{Preliminaries}
\label{sec:preliminaries}
The task aims to find the most corresponding position $(x,y)$ described by linguistic descriptions within the large-scale urban map $M$. To simplify this difficult task, we first divide the urban map into cubic submaps, denoted as $\{M_i \}_{i=1}^{m} $, where $m$ represents the number of submaps. Each submap $M_i=\{P_{i,j}|j=1,...,p\}$ consists of a set of object point clouds $P_{i,j} \in \mathbb{R}^{s\times3}$, where $s$ is the number of points in the object. Each object $P_{i,j}$ is obtained through semantic segmentation from the point cloud of the submap $M_i$. Besides, text query $T$ is represented as a set of hints $\{h_t \}_{t=1}^h$, where each hint describes the spatial relationships between the target position and its surrounding objects.


\begin{figure}[t]
    \centering
    \includegraphics[width=0.45\textwidth]{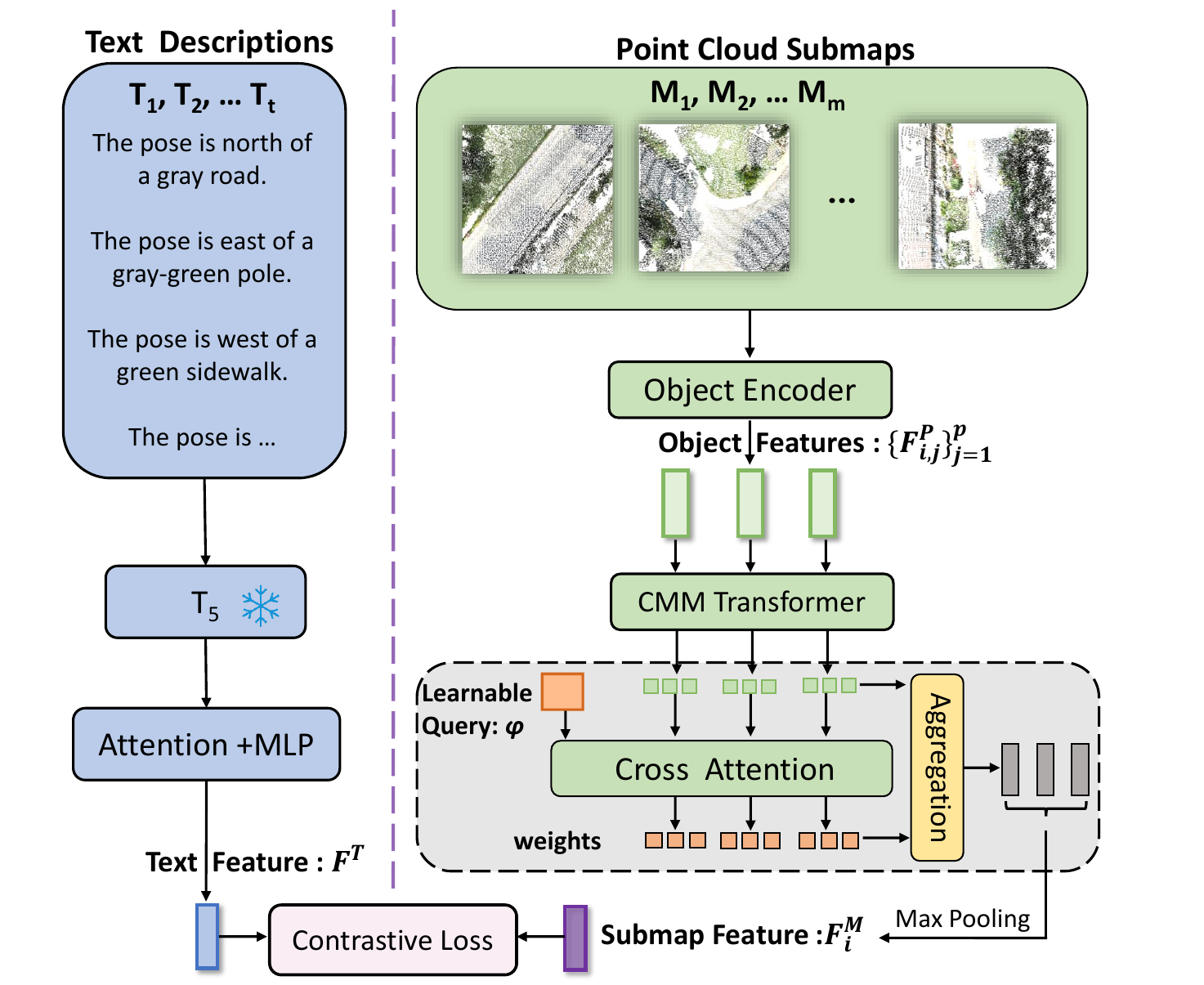}
    \caption{Illustration of coarse submap retrieval. We introduce the CMM Transformer and spatial consolidation scheme in the object encoding branch to model the partial relevance between 3D objects and achieve a better representation of the submap. Note that the T5 model in the text encoding branch is frozen during this process.}
    \label{fig:coarse}
    \vspace{-1.5em}
\end{figure}

\section{Method}
\label{sec:method}

We adopt a coarse-to-fine pipeline \cite{kolmet2022text2pos, xia2024text2loc} and propose a novel framework CMMLoc to address the language-based point cloud localization problem as shown in \cref{fig:overview}. In the coarse stage, the goal of the task is to identify the submap most likely to contain the target position. Recognizing the partially relevant nature of this task, we design a Cauchy-Mixture-Model-based Transformer and a spatial consolidation scheme for better representation of submaps, as detailed in \cref{sec:coarse}. In the fine stage, we focus on locating the target position based on $T$ and retrieved candidate submaps. We first pre-align text and object features via pre-training to bring 3D objects closer to the text modality, and then we introduce a cardinal direction integration module to capture the spatial relationships among objects for more fine-grained interactions between two modalities, described in \cref{sec:fine}. Training loss functions are described in \cref{sec:loss}.


\subsection{Coarse text-submap retrieval}
\label{sec:coarse}
Following \cite{kolmet2022text2pos,wang2023text,xia2024text2loc}, we employ a dual branch to learn global descriptors of text query $T$ and submap $M_i$ as is shown in \cref{fig:coarse}. Previous works solve this problem by matching the extracted global descriptors of text query and object point clouds, but they overlook the \textit{partially relevant characteristic} between the text descriptions and 3D objects in the submap. This oversight can degrade the global representation and lead to suboptimal submap retrieval performance. To address this, we formulate the coarse stage as a partially relevant retrieval problem and introduce a Cauchy-Mixture-Model-based Transformer in the submap encoding branch. Additionally, we propose a spatial consolidation scheme to improve submap representation.

\textbf{Text encoder.} Given a text query, we first extract its features using the frozen pre-trained model T5 ~\cite{raffel2020exploring}, which can help to get good initial embeddings due to the extensive pre-trained knowledge in T5. Next, we utilize the attention mechanism \cite{vaswani2017attention} to capture the contextual relationships between words and sentences. Specifically, we feed the text embeddings into the transformer block that involves multi-head self-attention and multi-layer perceptron (MLP) layers. This approach enables us to obtain better global descriptors of text query $F^T$, which pays attention to both the overall semantic meaning and the details.

\textbf{Object encoder.} To get the global descriptor of submap $M_i$, we adopt the approach that we encode every object $P_{i,j}$ in the submap and then aggregate the features to represent the submap. We first encode each object with PointNet++ \cite{qi2017pointnet++} to get the semantic embedding. However, semantic features alone are not sufficient, as each object also has distinctive properties like color, position, and point number. Therefore, we use separate encoders for color, position, and point number, each consisting of a 3-layer MLP with the same embedding dimension as the semantic embedding. We then concatenate the embeddings from all encoders to form the initial representation of each object in the submap.

\begin{figure}
    \centering
    \includegraphics[width=0.45\textwidth]{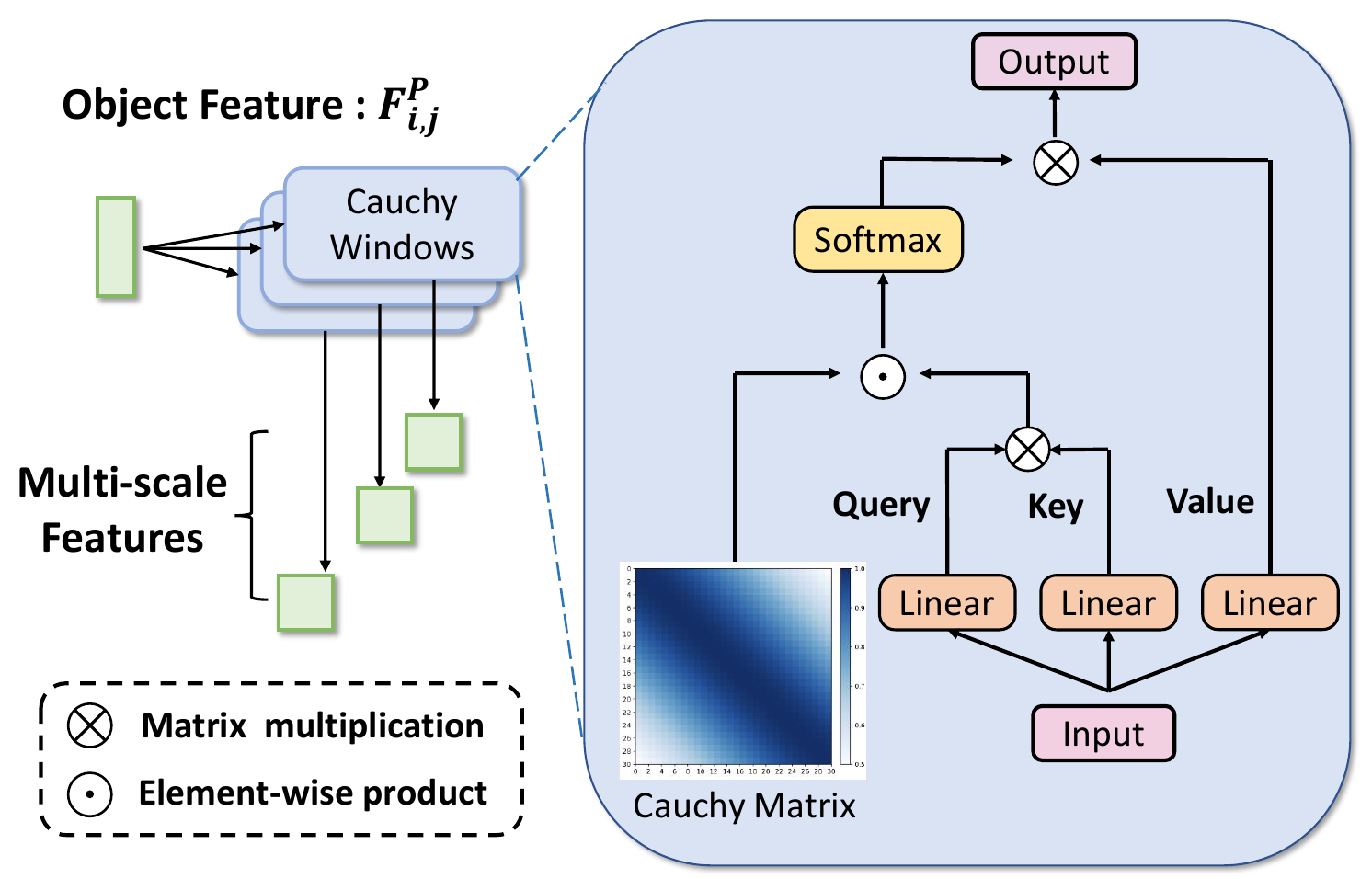}
    \caption{Illustration of CMM Transformer.}
    \label{fig:cmm}
    \vspace{-1.5em}
\end{figure}

\textbf{Cauchy-Mixture-Model-based Transformer.} Previous works typically use the output of the object encoder as the final object embedding and aggregate them to represent the submap, which neglects the partially relevant property. To better model the entire submap, we feed the object embeddings into our proposed Cauchy-Mixture-Model based Transformer (CMMT) inspired by \cite{gmmweight,wang2024gmmformer}. Specifically, CMMT is composed of several Cauchy windows, which incorporate Cauchy-Mixture-Model priors in the encoding process to encourage semantic similarity modeling as shown in \cref{fig:cmm}. Given object features $\{F^{P}_{i,j} \in \mathbb{R}^d\}_{j=1}^{p}$ in submap $M_i$, we first represent them as a matrix $X_i \in \mathbb{R}^{p\times d}$. In the Cauchy window, we project $X_i$ to query, key, value matrix $W^q$, $W^k$, $W^v$ through learnable parameters. We then use the query matrix to perform scaled dot-product attention over the key matrix, yielding an attention matrix score. Finally, we apply a Cauchy matrix $W^c \in \mathbb{R}^{p\times p}$ to perform element-wise product over the attention scores:
\begin{gather}
X_i^{attn} = \text{Softmax} (W^c \odot \frac{X_i W^q(X_i W^k)^{\top}}{\sqrt{d_k}})X_i W^v, \\
          W^c(i,j) = \frac{1}{\pi \gamma [1+(\frac{j-i}{\gamma})^2]},
\end{gather}
where $d_k$ is the dimension of queries and keys, $\gamma$ is the scale parameter of Cauchy density distribution and $\odot$ indicates the element-wise product function. After obtaining $X_i^{attn}$, we pass it through a Feed-Forward Network (FFN) to get the output of the Cauchy window $X_i^{output}$. We have $N$ parallel Cauchy windows with different scales as objects in every submap are various. 

The Cauchy window is designed to effectively model local relationships among 3D objects by assigning different Cauchy weights within the feature matrix $X_i$. In other words, the order of object features is significant as adjacent features receive higher Cauchy attention weights. We explore two approaches for Cauchy matrix values assignment: one based on physical spatial distance and the other based on semantic similarity. Distance-based assignment means objects positioned closer get higher Cauchy weights, and assignment by semantic similarity means greater semantic similarity yields higher weights. Experiments in \cref{sec:albation} prove that the distance-based approach is less effective. Therefore, we adopt semantic similarity for weight assignment. To implement this, objects with the same semantic label are grouped together during the semantic segmentation, and the order of labels is randomized.
Although random sorting is not optimal, we find that it yields comparable performance due to the heavy-tailed properties of the Cauchy distribution. Unlike Gaussian windows, the Cauchy window assigns higher probabilities to objects that are farther apart, making it more robust to outliers and better suited for modeling phenomena with unpredictable elements. A comparative analysis of the Cauchy and Gaussian distributions is presented in \cref{sec:albation}  
\begin{table*}[t]
    \centering
    \small
    \begin{tblr}{
      cells = {c},
      cell{1}{2} = {c=6}{},
      cell{2}{2} = {c=3}{},
      cell{2}{5} = {c=3}{},
      hline{1,8} = {-}{0.08em},
      hline{2-3} = {2-7}{0.03em},
      hline{4,7} = {-}{0.05em},
    }
                    & Localization Recall ($\epsilon < 5/10/15m $) $\uparrow$ &                                           &                                           &                                           &                                           &                                           \\
    Methods         & Validation Set                                          &                                           &                                           & Test Set                                  &                                           &                                           \\
                    & $k=1$                                                   & $k=5$                                     & $k=10$                                    & $k=1$                                     & $k=5$                                     & $k=10$                                    \\
    Text2Pos~\cite{kolmet2022text2pos}       & 0.14/0.25/0.31                                          & 0.36/0.55/0.61                            & 0.48/0.68/0.74                            & 0.13/0.21/0.25                            & 0.33/0.48/0.52                            & 0.43/0.61/0.65                            \\
    RET~\cite{wang2023text}           & 0.19/0.30/0.37                                          & 0.44/0.62/0.67                            & 0.52/0.72/0.78                            & 0.16/0.25/0.29                            & 0.35/0.51/0.56                            & 0.46/0.65/0.71                            \\
    Text2Loc~\cite{xia2024text2loc}       & 0.37/0.57/0.63                                          & 0.68/0.85/0.87                            & 0.77/0.91/0.93                            & 0.33/0.48/0.52                            & 0.61/0.75/0.78                            & 0.71/0.84/0.86                            \\
    CMMLoc (Ours) & \textbf{0.44}/\textbf{0.62}/\textbf{0.68}               & \textbf{0.75}/\textbf{0.88}/\textbf{0.90} & \textbf{0.83}/\textbf{0.93}/\textbf{0.95} & \textbf{0.39}/\textbf{0.53}/\textbf{0.56} & \textbf{0.67}/\textbf{0.80}/\textbf{0.82} & \textbf{0.77}/\textbf{0.87}/\textbf{0.89} 
    \end{tblr}
    \caption{Performance comparison on the KITTI360Pose benchmark~\cite{kolmet2022text2pos}.}
    \label{tab:fine_results}
    \vspace{-1em}
\end{table*}


Notably, in practical scenarios, misclassification can occur during the semantic segmentation process,  which can affect the performance of our CMMLoc, as it relies on the semantic labels of the objects. To evaluate the robustness of CMMLoc under such conditions, we conduct an ablation study examining its performance under varying misclassification rates, as detailed in \cref{sec:albation}.
\textbf{Spatial Consolidation Scheme.} Considering the irregularity of 
the point cloud, we design a learnable query $\varphi$ to learn adaptive aggregation weights for aggregating object features with varying receptive fields. In particular, we use a cross-attention layer to generate aggregation weights: 
\begin{gather}
    w_n = \text{Linear}(\text{Softmax}(\frac{\varphi X_{i,n}^{output}}{\sqrt{d_k}})X_{i,n}^{output}),\\  
    \tilde X_i^{output} =  \sum_{n=1}^N w_{n} X_{i,n}^{output},
\end{gather}
where $n$ represents the index of the Cauchy window. After obtaining the object features in the submap, we apply a max pooling operation 
to obtain the final global descriptor of submap $F^{M}_i$ for the partially relevant retrieval.  
 

\begin{figure}[t]
    \centering
    \includegraphics[width=0.48\textwidth]{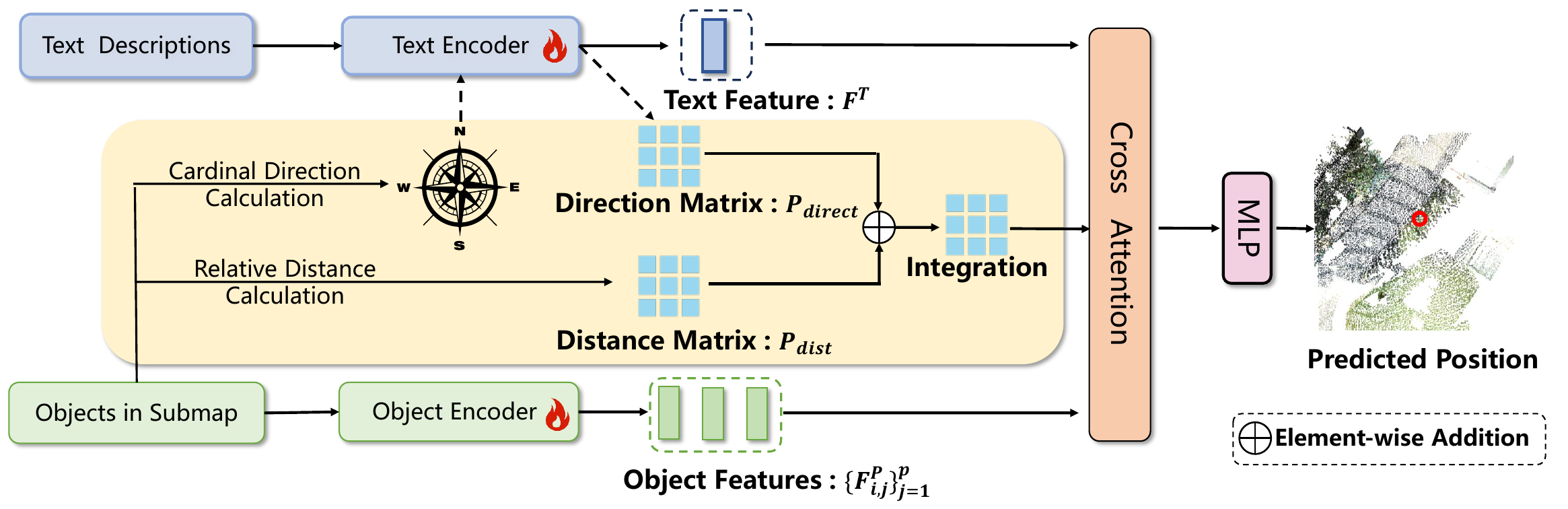}
    \caption{Illustration of Cardinal Direction Integration.}
    \label{fig:cdi}
    \vspace{-1em}
\end{figure}
\subsection{Fine localization}
\label{sec:fine}
Following \cite{xia2024text2loc}, we adopt a matching-free network as the base model for fine localization. Unlike the coarse stage, the key challenge in the fine stage is to encourage fine-grained interactions between the text queries and 3D objects in the submap. To this end, we propose a strategy: we first pre-align the text encoder and object encoder to bring 3D object features closer to the text modality. Then we introduce a Cardinal Direction Integration method to capture the spatial relations between objects in the submap as a supplement for common positional embedding.

\textbf{Pre-alignment.} The main challenge in this task lies in the huge gap between the text modality and object point clouds, which hinders fine-grained alignment. To bridge this gap and make 3D objects more “language-like”, we pre-train both the text encoder and object encoder before training the model for localization refinement. Since the object labels and colors are the most relevant textual information for 3D objects, we involve the color encoder, the object encoder, and the text encoder for pre-alignment. Specifically, the color encoder and object encoder extract color embeddings $F^P_{color}$ and object features $F^P_{object}$ from the object point cloud, while the text encoder encodes the color and label information in the text query to produce text embeddings $F^T_{color}$ and $F^T_{label}$. To provide a good initialization for the fine stage and facilitate better alignment, we aim to minimize the distance of matched embeddings, as is outlined in the loss function in \cref{sec:loss}.



\textbf{Cardinal Direction Integration.} The position encoder in the object encoder extracts the absolute position embedding of objects within the submap,  but we believe this is insufficient for the fine-grained alignment between texts and objects. To address this, we introduce a Cardinal Direction Integration (CDI) module to capture the spatial relations between objects as illustrated in \cref{fig:cdi}. Specifically, we compute the pair-wise distances between the center coordinates of objects in the submap, resulting in a distance matrix $P_{dist}\in \mathbb{R}^{p\times p}$. Additionally, we determine the relative cardinal direction between pairs of objects. For example, if object A is situated to the east of object B, the cardinal direction ``east" is extracted. This textual directional information ``east" is then encoded by the text encoder and fed into an MLP to form the direction matrix $P_{direct}\in \mathbb{R}^{p\times p}$. We combine both the distance and direction matrices to form the complete relative position matrix $P\in \mathbb{R}^{p \times p}$:
\begin{gather}
 P = P_{direct} + \alpha * P_{dist},
\end{gather}  
where $\alpha$ is a weight factor. Then we add the relative position matrix to the computed attention weights before doing the softmax in the attention layer. Considering the input for an attention layer: queries $Q$ of size $p\times d_f$, keys $K$ of size $p\times d_f$ and values V and the relative position matrix $P$, the output of the attention layer is:
\begin{gather}
    A=\frac{QK^{\top}+P}{\sqrt{d_f}}.
\end{gather}
Finally, we obtain the predicted position by passing the output of the attention layer through an MLP. 
\subsection{Loss Functions}
\label{sec:loss}
For the coarse stage, given a batch of global descriptors of submaps $\{ F^M_i \}_{i=1}^{B} $ and text queries $\{F^T_i \}_{i=1}^B $, where $B$ represents the batch size, we employ contrastive loss between each pair to train our model instead of the pairwise ranking loss in previous work:
\begin{footnotesize}
\begin{gather}
      l(i,T,M) = -\log\frac{\exp(F^T_i\cdot F^M_{i}/\tau)}{\sum\limits_{j\in N} \exp(F_i^T\cdot F_j^M/\tau)} - \log\frac{\exp(F^M_i\cdot F^T_{i}/\tau)}{\sum\limits_{j\in N} \exp(F_i^M\cdot F_j^T/\tau)},       
\end{gather}
\end{footnotesize}where $\tau$ represents the temperature coefficient. Within a training batch, the final contrastive loss is computed by summing $l(i,T,M)$ and dividing by the batch size.  

As for the fine stage, the objective is to minimize the difference between the predicted position and the ground truth. We take the strategy which includes pre-training first and then incorporate Cardinal Direction Integration for more fine-grained alignment. Specifically, we first use mean squared error loss for pre-alignment to make 3D objects more ``like'' language:
\begin{equation}
    L_{pre} = \big \|F_{color}^P - {F}_{color}^T   \big \|_{2}+\big \|F_{object}^P - {F}_{label}^T   \big \|_{2},
\end{equation}
then we continue to use the mean squared error loss to train the translation regressor:
\begin{equation}
    L(P_{gt},P_{pred})=\big \|P_{gt} - {P}_{pred}   \big \|_{2},
\end{equation}
where $P_{pred}=(x,y)$ is the coordinate of predicted position, and $P_{gt}$ is the ground truth coordinates.

\begin{table}[t]
    \centering
    \resizebox{0.46\textwidth}{!}{
    \begin{tblr}{
      cells = {c},
      cell{1}{2} = {c=6}{},
      cell{2}{2} = {c=3}{},
      cell{2}{5} = {c=3}{},
      hline{1,8} = {-}{0.08em},
      hline{2-3} = {2-7}{0.03em},
      hline{4,7} = {-}{0.05em},
    }
                    & Submap Retrieval Recall $\uparrow$ &               &               &               &               &               \\
    Methods          & Validation Set                     &               &               & Test Set      &               &               \\
                    & $k=1$                              & $k=3$         & $k=5$         & $k=1$         & $k=3$         & $k=5$         \\
    Text2Pos~\cite{kolmet2022text2pos}       & 0.14                               & 0.28          & 0.37          & 0.12          & 0.25          & 0.33          \\
    RET~\cite{wang2023text}            & 0.18                               & 0.34          & 0.44          & -             & -             & -             \\
    Text2Loc~\cite{xia2024text2loc}            & 0.32                               & 0.56          & 0.67          & 0.28             & 0.49             & 0.58             \\
    CMMLoc (Ours) & \textbf{0.35}                      & \textbf{0.61} & \textbf{0.73} & \textbf{0.32} & \textbf{0.53} & \textbf{0.63} 
    \end{tblr}}
    \caption{Performance comparison for coarse text-submap retrieval on the KITTI360Pose benchmark~\cite{kolmet2022text2pos}. Note that only values that are available in RET~\cite{wang2023text} are reported.}
    \label{tab:coarse_results}
    \vspace{-1em}
\end{table}

\section{Experiments}
\label{sec:experiment}
\subsection{Dataset \& Evaluation metrics}
We conduct our experiments including training and evaluating on the KITTI360Pose dataset \cite{kolmet2022text2pos}, which is built upon the KITTIPose dataset by \cite{liao2022kitti} with sampled locations and generated hint descriptions. It contains point clouds of 9 urban scenes, covering 14,934 positions and 43,381 position-query pairs with an area of 15.51 $\text{km}^2$ in all. We follow the setting in \cite{kolmet2022text2pos} and choose 5 scenes for training, 1 scene for validation, and 3 scenes for testing. Each submap is a 3D cube with a fixed size of 30m and a stride of 10m. The whole dataset has 11,259/1,434/4,308 submaps for training/validation/testing scenes and 17,001 submaps in all.
\begin{table}[t]
    \centering
    \small
    \resizebox{0.47\textwidth}{!}{
    \begin{tblr}{
      cells = {c},
      cell{1}{1} = {r=3}{},
      cell{1}{2} = {c=6}{},
      cell{2}{2} = {c=3}{},
      cell{2}{5} = {c=3}{},
      hline{1,8} = {-}{0.08em},
      hline{2-3} = {2-7}{0.03em},
      hline{4,7} = {-}{0.05em},
    }
    Methods      & Submap Retrieval Recall $\uparrow$ &               &               &               &               &               \\
                & Validation Set                     &               &               & Test Set      &               &               \\
                & $k=1$                              & $k=3$         & $k=5$         & $k=1$         & $k=3$         & $k=5$         \\
    Transformer~\cite{xia2024text2loc}      & 0.32                               & 0.56          & 0.67          & 0.28          & 0.49          & 0.58          \\
    GMMFormer~\cite{wang2024gmmformer}     & 0.33                               & 0.57          & 0.68          & 0.30 & 0.50          & 0.60          \\
    CMMT & 0.33                    & 0.58 & 0.69 & 0.31 & 0.52 & 0.62 \\
    CMMT-SC & \textbf{0.35}                      & \textbf{0.61} & \textbf{0.73} & \textbf{0.32} & \textbf{0.53} & \textbf{0.63}
    \end{tblr}}
    \caption{Ablation study of coarse text-submap retrieval on the KITTI360Pose benchmark \cite{kolmet2022text2pos}. Transformer indicates the attention mechanism in \cite{xia2024text2loc}. GMMFormer indicates the Gaussian-Mixture-Model-based Transformer designed for partially relevant video retrieval in \cite{wang2024gmmformer}. CMMT denotes the Cauchy-Mixture-Model-based Transformer we propose. CMMT-SC refers to CMMT with the spatial consolidation scheme.}
    \label{tab:coarse ablation study}
    \vspace{-1.0em}
\end{table}

\begin{figure}
    \centering
    \includegraphics[width=0.45\textwidth]{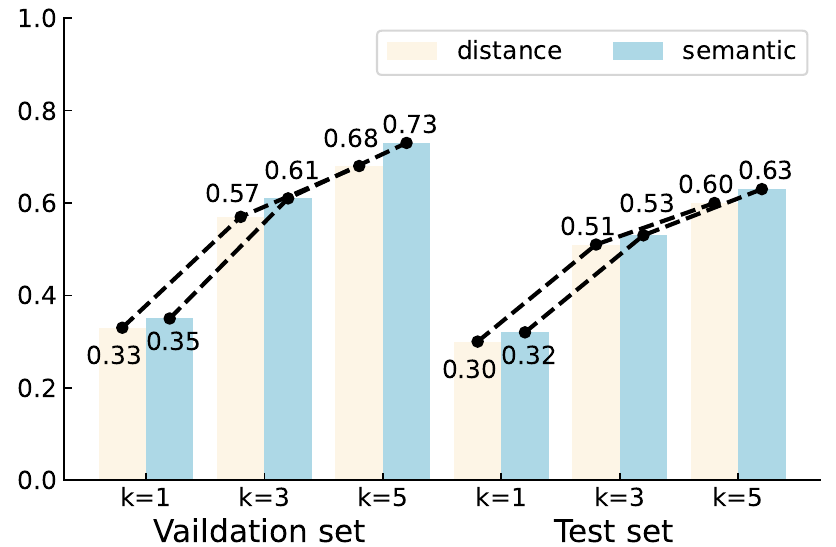}
    \caption{Performance comparison of coarse submap retrieval with different CMM weight allocation strategies on the KITTI360Pose dataset \cite{kolmet2022text2pos}.}
    \label{fig:dist-semantic}
    \vspace{-1.5em}
\end{figure} 
We formulate the coarse stage as a partially relevant retrieval problem and evaluate its performance using retrieval recall at Top $k$ $(k \in \{1,3,5\} )$. For fine localization, we utilize localization recall to assess localization capability. Specifically, based on the Top k retrieval candidates $(k\in \{ 1,5,10\})$, we measure the ratio of successfully localized queries if the error is below specific error thresholds, namely, $\epsilon$ $<$ 5/10/15m by default.


\subsection{Comparison with State-of-the-art Methods}
Compared to existing methods, our method achieves the best performance, demonstrating its superiority. We evaluate the performance of our model on the KITTI360Pose validation and test sets for a fair comparison. For coarse submap retrieval, we report the results in \cref{tab:coarse_results}. As shown, our model significantly outperforms previous methods. On the validation set, our best performance achieves a Top-1 recall of 0.35, exceeding the current state-of-the-art Text2Loc by 9\%. On the test set, recall rates achieved at Top-1, Top-3, and Top-5 are 0.32, 0.53, and 0.63, respectively, improving on Text2Loc by 14\%, 9\%, and 8\%. For fine localization, the results are shown in \cref{tab:fine_results}. We achieve a Top-1 recall rate of 0.44 on the validation set and 0.39 on the test set under an error bound of $\epsilon <$ 5m, which are 19\% and 18\% higher than Text2Loc, respectively. Our method also maintains superior performance when relaxing the localization error constraints or increasing $k$.

\begin{table}[t]
    \centering
    \small
    \resizebox{0.47\textwidth}{!}{%
    \begin{tblr}{
      cells = {c},
      cell{1}{2} = {c=6}{},
      cell{2}{2} = {c=3}{},
      cell{2}{5} = {c=3}{},
      hline{1,9} = {-}{0.08em},
      hline{2-3} = {2-7}{0.03em},
      hline{4,8} = {-}{0.05em},
    }
                    & Localization Recall ($\epsilon < 5m$) $\uparrow$ &               &               &               &               &               \\
    Methods         & Validation Set                                   &               &               & Test Set      &               &               \\
                    & $k=1$                                            & $k=5$         & $k=10$        & $k=1$         & $k=5$         & $k=10$        \\
    Text2Loc~\cite{xia2024text2loc}       & 0.37                                             & 0.68          & 0.77          & 0.33          & 0.61          & 0.71          \\                    
    Text2Loc*       & 0.42                                             & 0.74          & \textbf{0.83}          & 0.37          & 0.65          & 0.74          \\
    CMMLoc\_PA   & 0.43                                             & 0.74          & \textbf{0.83}          & 0.38          & 0.66          & 0.75          \\
    CMMLoc\_CDI    & 0.43                                             & 0.73          & 0.81          & 0.38          & 0.66          & 0.76          \\
    CMMLoc (Ours) & \textbf{0.44}                                    & \textbf{0.75} & \textbf{0.83} & \textbf{0.39} & \textbf{0.67} & \textbf{0.77} 
    \end{tblr}}
    \caption{Ablation study of fine localization on the KITTI360Pose benchmark. Text2Loc* indicates the fine localization network from Text2Loc, with submaps retrieved through our coarse model. CMMLoc\_PA indicates the removal of the CDI while retaining the Pre-alignment process in our network. Conversely, CMMLoc\_CDI keeps the CDI but removes the Pre-alignment.}
    \label{tab: fine ablation study}
    \vspace{-1.5em}
\end{table}
\subsection{Ablation study}
\label{sec:albation}
In this section, we conduct ablation studies to evaluate the effectiveness of the proposed modules in CMMLoc for coarse submap retrieval and fine localization stages.

\textbf{Coarse text-submap retrieval.} For the coarse stage, our focus is on assessing the contribution of the proposed Cauchy-Mixture-Model-based Transformer and spatial consolidation scheme. As we are the first to formulate the sub-task in the coarse stage as a partially relevant retrieval problem, we compare CMMT with GMMFormer proposed in \cite{wang2024gmmformer} and Transformer architecture used in Text2Loc \cite{xia2024text2loc}. As is illustrated in \cref{tab:coarse ablation study}, GMMFormer \cite{wang2024gmmformer}, which was originally designed for partially relevant video retrieval, outperforms the Transformer architecture. This demonstrates the necessity of modeling the coarse stage as a partially relevant retrieval problem from a different perspective. 
Benefiting from the robust nature of the Cauchy distribution, CMMT is more resilient to outliers in the submap, contributing to better performance compared to GMMFormer. With the spatial consolidation scheme, CMMT becomes more adaptive to different 3D shapes and achieves superior results. In comparison with the Transformer used in Text2Loc, CMMT-SC achieves 0.35 at Top-1 on the validation set and 0.32 at Top-1 on the test set, outperforming the Transformer by margins of 9\% and 14\%, respectively. Based on the above analysis, we demonstrate the correctness of modeling the coarse stage as a partially relevant retrieval problem and the superiority of the proposed Cauchy-Mixture-Model-based Transformer and spatial consolidation scheme.

\begin{figure}
    \centering
    \includegraphics[width=0.49\textwidth]{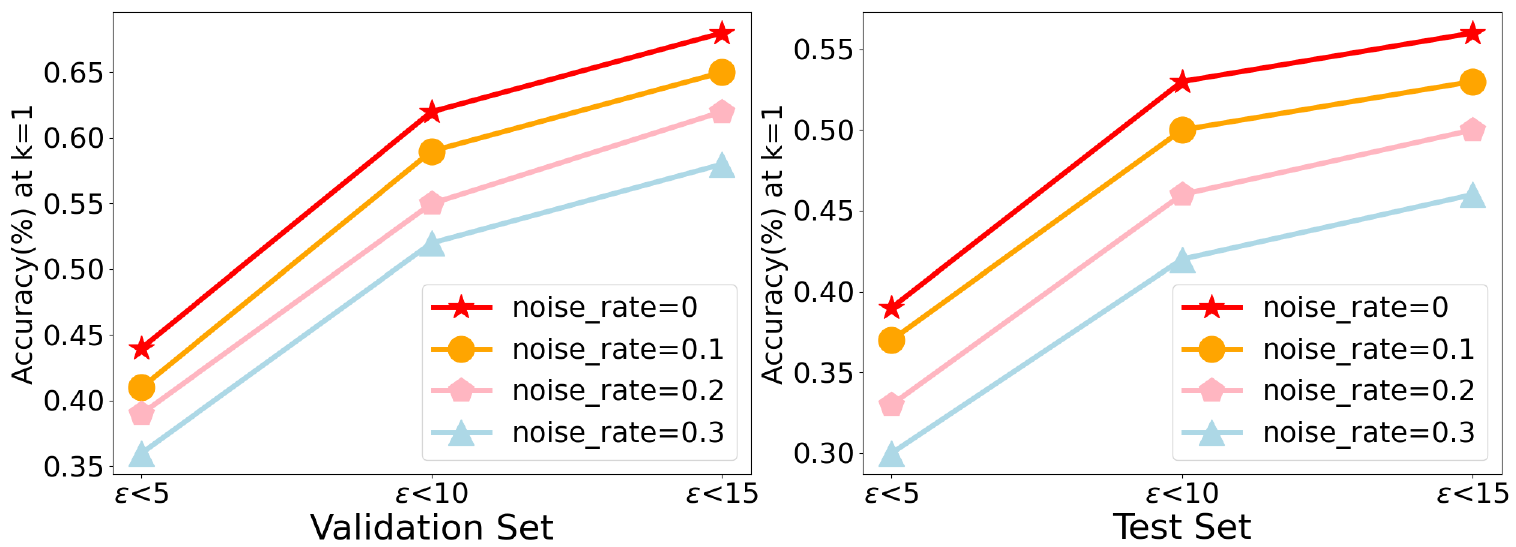}
    \caption{Performance Comparison at varying noise rates of the predicted object labels on the validation and test sets.}
    \label{fig:noise}
    \vspace{-1.5em}
\end{figure} 

\begin{figure*}[t]
    \vspace{-0.7em}
    \centering
    \includegraphics[width=0.95\textwidth]{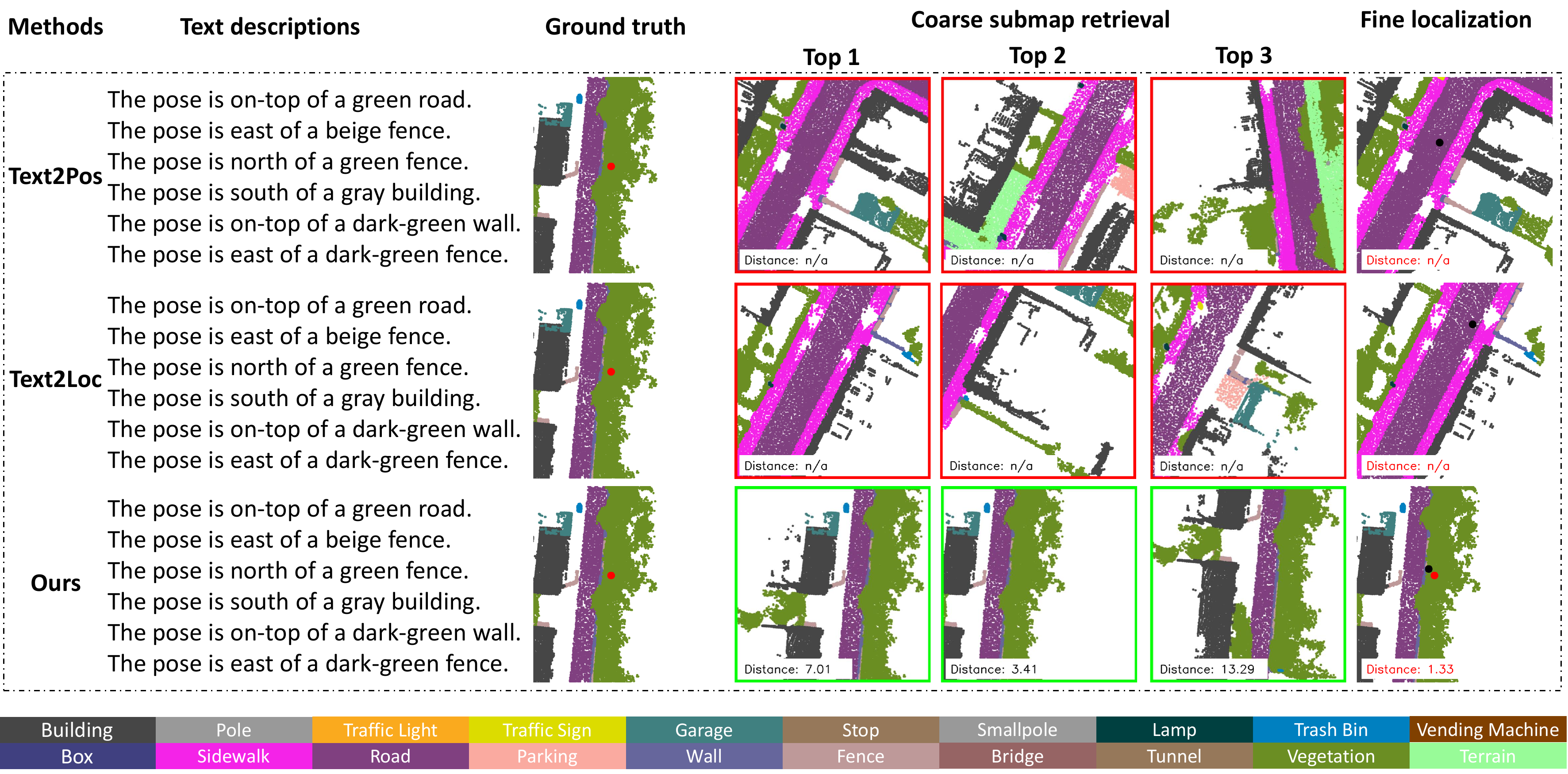}
    \caption{ Qualitative localization results on the KITTI360Pose dataset: In coarse submap retrieval, green boxes indicate positive submaps containing the target location, while red boxes represent negative submaps. For fine localization, red and black dots correspond to the ground truth and predicted target locations. We label the distances between the prediction and ground truth in the submap, with ``n/a" indicating submaps from different scenes.}
    \label{fig:rgb}
    \vspace{-1em}
\end{figure*}

\begin{figure}
    \centering
    \includegraphics[width=0.48\textwidth]{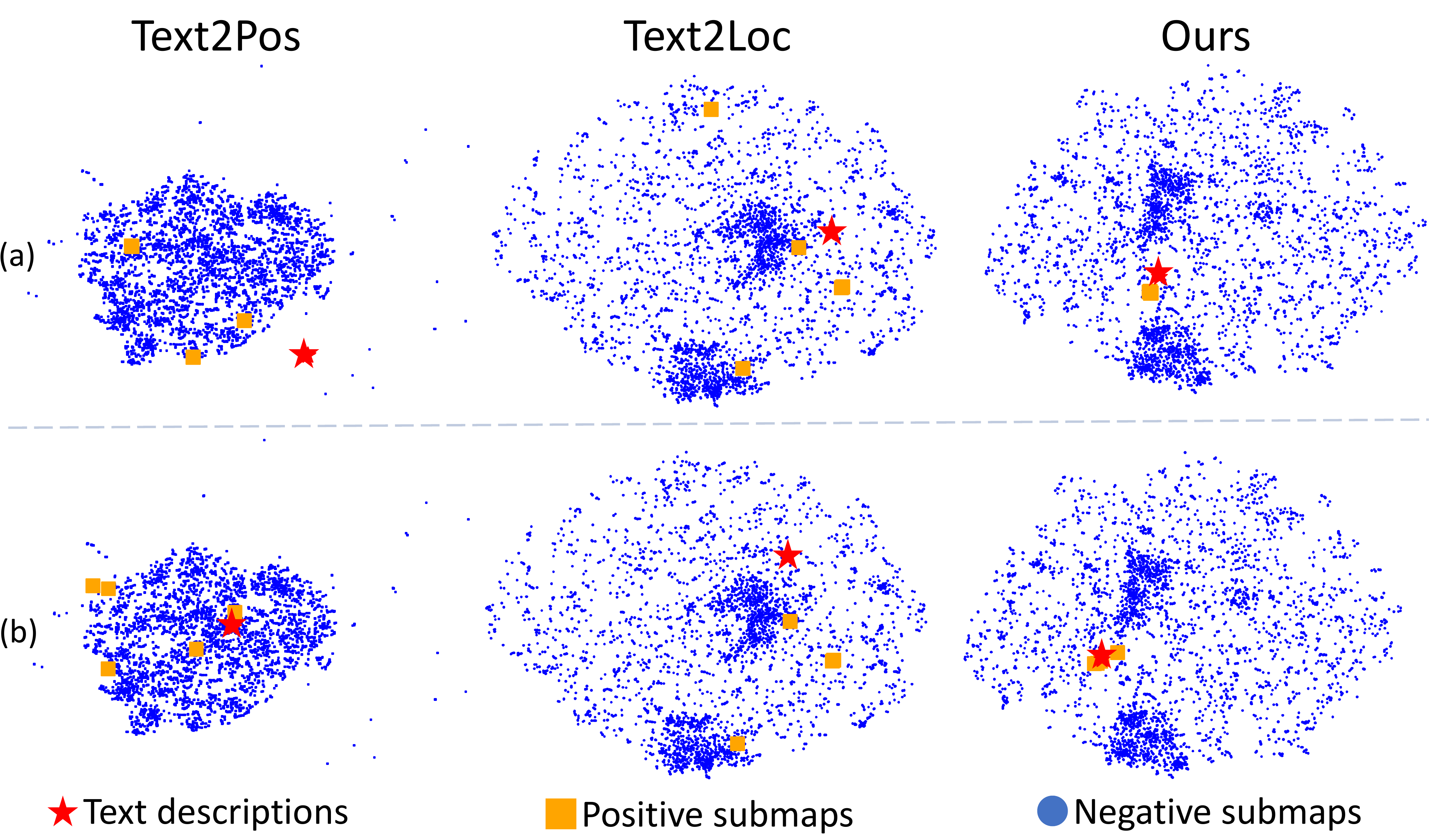}
    \caption{T-SNE visualization for coarse submap retrieval.}
    \label{fig:point}
    \vspace{-1.5em}
\end{figure}

Moreover, we explore two different Cauchy weight assignment methods: distance-based and semantic similarity-based. The performance comparison on the KITTI360Pose dataset is shown in \cref{fig:dist-semantic}. As shown, assigning Cauchy weights based on semantic similarity performs better on both the validation and test sets, demonstrating the effectiveness of weighting according to semantic similarity. Furthermore, we consider practical scenarios where semantic labels are not readily available and require the pre-processing step of semantic segmentation. Under this circumstance, we investigate the impact of incorrect semantic labels on the performance of our model. Specifically, we add noise to the labels of the validation and test sets to simulate errors in semantic segmentation and then use the noisy dataset to evaluate the performance of our model. As is shown in \cref{fig:noise}, the performance of CMMLoc decreases as the noise rate increases, demonstrating its reliance on semantic labels. Notably, at a noise rate of 0.1, CMMLoc still outperforms Text2Loc \cite{xia2024text2loc}, and at a noise rate of 0.2, its performance remains comparable, indicating the robustness of our method.

\textbf{Fine localization.} To validate the effectiveness of the Pre-alignment and Cardinal Direction Integration Module in the fine localization stage, we separately conduct ablation studies on them, denoted as CMMLoc\_PA and CMMLoc\_CDI. To fairly assess the results, we use the submaps retrieved from our coarse stage and report the results in \cref{tab: fine ablation study}. As shown, Text2Loc* significantly outperforms the origin results of Text2Loc, indicating the superiority of modeling the coarse stage as a partially relevant retrieval problem, as well as the proposed CMMT and spatial consolidation scheme. Moreover, CMMLoc\_PA and CMMLoc\_CDI both have better performance than Text2Loc*, achieving about a 15\% improvement at Top-1 on the test set, showing the effectiveness of our proposed modules.

\textbf{Visualization.} We compare our learned embedding space with previous works by T-SNE \cite{van2008visualizing} in \cref{fig:point}. As is shown, Text2Pos \cite{kolmet2022text2pos} and Text2Loc \cite{xia2024text2loc} generate less discriminative embedding space, with positive submaps scattered and distant from the query descriptions. In contrast, our method effectively aligns positive submaps with text queries, yielding a more discriminative space for coarse submap retrieval. Moreover, we present some qualitative results in \cref{fig:rgb}. Given the same text query, we visualize the ground truth, the Top-3 retrieved submaps, and the fine localization results of previous work and ours. Submaps containing the target position are defined as positive. CMMLoc demonstrates superior performance in submap retrieval and localization refinement, not only identifying the positive submaps but also achieving more accurate localization.

\section{Conclusion}
\label{conclusion}
We analyze existing works on language-based 3D point cloud localization and identify the limitations of the current coarse-to-fine pipeline, namely the neglect of partial relevance between text descriptions and 3D objects and the failure to model this uncertainty. We propose CMMLoc to tackle this problem. In the coarse stage, we are the first to model the task as a partially relevant retrieval problem and propose a Cauchy-Mixture-Model-based Transformer and a spatial consolidation scheme for local modeling of 3D objects and better submap representation. In the fine stage, we enhance fine-grained interactions between texts and objects by pre-aligning text queries with 3D objects and applying a cardinal direction integration module to capture spatial relations between objects. Extensive experiments show that our CMMLoc significantly outperforms the current state-of-the-art, highlighting its effectiveness.
\section{Acknowledgements}

This research was also supported by the advanced computing resources provided by the Supercomputing Center of the USTC. We also acknowledge the support of GPU cluster built by MCC Lab of Information Science and Technology Institution, USTC.

{
    \small
    \bibliographystyle{ieeenat_fullname}
    \bibliography{main}
}

\clearpage
\setcounter{page}{1}
\maketitlesupplementary
\setcounter{section}{0}
\renewcommand\thesection{\Alph{section}}
\section{Overview}

In this supplementary material, we provide additional experiments and visualization results to further demonstrate the effectiveness of our proposed CMMLoc. In \cref{sec:ablation_cmmt}, we conduct an ablation study on our proposed Cauchy-Mixture-Model-based Transformer with different layer numbers on the KITTI360Pose dataset \cite{kolmet2022text2pos}, covering both the coarse and fine stages. Moreover, we present the implementation details of our model in \cref{sec:details} and more visualization results in \cref{sec:visual}.

\section{More analysis of Cauchy-Mixture-Model-based Transformer}
\label{sec:ablation_cmmt}
In this section, we primarily analyze the impact of different layer numbers of the proposed Cauchy-Mixture-Model-based Transformer with spatial consolidation scheme on the performance of our CMMLoc, including the coarse submap retrieval stage and the fine localization stage on the KITTI360Pose dataset \cite{kolmet2022text2pos}.

For the retrieval of coarse submap, \cref{tab:coarse_number} shows the performance of CMMLoc with different numbers of CMMT-SC, where `0' means that we use the vanilla attention mechanism in Text2Loc \cite{xia2024text2loc} instead of our proposed CMMT-SC. As shown, CMMLoc achieves the best performance with a Cauchy-Mixture-Model-based Transformer with spatial consolidation scheme. When the number is set to 2, the performance drops significantly, falling below that of Text2Loc. The performance in fine localization exhibits a similar trend. As shown in \cref{tab: fine_number}, CMMLoc performs best with a single CMMT-SC, but its performance declines sharply when the number is increased to 2.

The possible explanation for the performance degradation when using 2 layers of CMMT-SC is that the heavy-tailed nature of the Cauchy distribution enhances robustness to outliers during modeling. However, applying it consecutively may excessively blur the differences between the original features, thereby reducing the discriminative capability of the model. As a result, we set the fixed number of CMMT-SC as 1 in our CMMLoc. 

\section{Implementation Details}
\label{sec:details}
\begin{table}[t]
    \centering
    \small
    \resizebox{0.47\textwidth}{!}{
    \begin{tblr}{
      cells = {c},
      cell{1}{1} = {r=3}{},
      cell{1}{2} = {c=6}{},
      cell{2}{2} = {c=3}{},
      cell{2}{5} = {c=3}{},
      hline{1,7} = {-}{0.08em},
      hline{2-3} = {2-7}{0.03em},
      hline{4} = {-}{0.05em},
    }
    Number of CMMT-SC      & Submap Retrieval Recall $\uparrow$ &               &               &               &               &               \\
                & Validation Set                     &               &               & Test Set      &               &               \\
                & $k=1$                              & $k=3$         & $k=5$         & $k=1$         & $k=3$         & $k=5$         \\
    0     & 0.32                               & 0.56          & 0.67          & 0.28          & 0.49          & 0.58          \\
    1    & \textbf{0.35}                               & \textbf{0.61}          & \textbf{0.73}          & \textbf{0.32} & \textbf{0.53}          & \textbf{0.63}          \\
    2 & 0.31                    & 0.55 & 0.66 & 0.27 &0.48 & 0.58 \\
    \end{tblr}}
    \caption{Coarse submap retrieval performance for CMMLoc with different numbers of CMMT-SC on the KITTI360Pose benchmark. CMMT-SC denotes the Cauchy-Mixture-Model-based Transformer with spatial consolidation scheme. `0' means using the vaniila attention architecture in Text2Loc \cite{xia2024text2loc}.}
    \label{tab:coarse_number}
\end{table}
We conduct our experiments on an NVIDIA A800 GPU. For the coarse submap retrieval, we train the model with Adam optimizer with a learning rate of 5e-4 for 20 epochs. We set the batch size to 64 and utilize a multi-step training schedule wherein the learning rate is decayed by 0.4 every 7 epochs. The temperature coefficient $\tau$ in the contrastive loss function is set to 0.1. Each submap contains a maximum of 28 objects. We employ PointNet++ \cite{qi2017pointnet++} in \cite{kolmet2022text2pos} to extract the semantic feature of every object in the submap, followed by a single Cauchy-Mixture-Model-based Transformer with spatial consolidation scheme for local modeling to achieve better submap representation. For the fine localization, we first pre-train the text encoder and object encoder with a learning rate of 3e-4 with batch size 32 to get a well-initialized state for localization refinement. Then we train the fine localization network with the same learning rate for 45 epochs. For a fair comparison, we set the embedding dimension to 256 for both the text and submap branches in coarse submap retrieval and 128 in fine localization, following the same configuration as in previous work.

\begin{table}[t]
    \centering
    \small
    \resizebox{0.47\textwidth}{!}{%
    \begin{tblr}{
      cells = {c},
      cell{1}{2} = {c=6}{},
      cell{2}{2} = {c=3}{},
      cell{2}{5} = {c=3}{},
      hline{1,7} = {-}{0.08em},
      hline{2-3} = {2-7}{0.03em},
      hline{4} = {-}{0.05em},
    }
                    & Localization Recall ($\epsilon < 5m$) $\uparrow$ &               &               &               &               &               \\
    Number of CMMT-SC         & Validation Set                                   &               &               & Test Set      &               &               \\
                    & $k=1$                                            & $k=5$         & $k=10$        & $k=1$         & $k=5$         & $k=10$        \\
  0      & 0.38                                             & 0.69          & 0.80          & 0.35          & 0.63          & 0.73          \\                    
    1& \textbf{0.44}                                    & \textbf{0.75} & \textbf{0.83} & \textbf{0.39} & \textbf{0.67} & \textbf{0.77} \\
     2       & 0.39                                             & 0.70          & 0.80          & 0.35          & 0.64          & 0.74          
    \end{tblr}}
    \caption{Localization performance for Text2Loc with different numbers of CMMT-SC on the KITTI360Pose benchmark. `0' means using the fine localization network from CMMLoc, with submaps retrieved through the coarse submap retrieval framework from Text2Loc \cite{xia2024text2loc}.}
    \label{tab: fine_number}
    \vspace{-1.0em}
\end{table}
\begin{figure*}[htp]
    \vspace{-0.7em}
    \centering
    \includegraphics[width=0.95\textwidth]{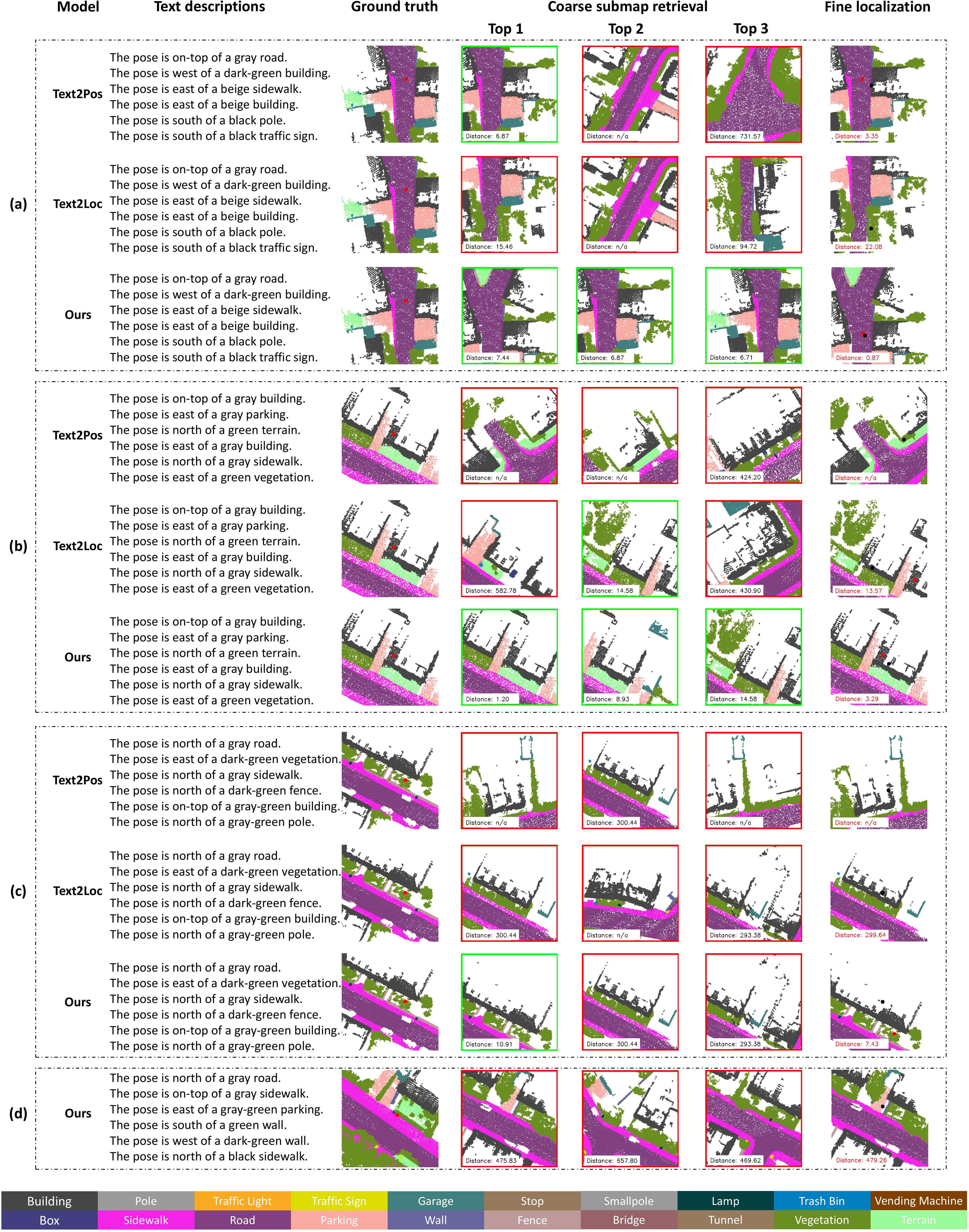}
    \caption{ Qualitative localization results on the KITTI360Pose dataset: In coarse submap retrieval, green boxes indicate positive submaps containing the target location, while red boxes represent negative submaps. For fine localization, red and black dots correspond to the ground truth and predicted target locations. We label the distances between the prediction and ground truth in the submap, with ``n/a" indicating submaps from different scenes.}
    \label{fig:supp}
    \vspace{-1em}
\end{figure*}
\section{More Visualization results}
\label{sec:visual}
In this section, we present additional visualizations to compare our CMMLoc with previous methods, as shown in \cref{fig:supp}, including an analysis of a failure case. For (a) and (b), CMMLoc successfully retrieves all positive submaps within the top-3 results during coarse submap retrieval, whereas most retrievals from Text2Loc and Text2Pos are incorrect. When the positive submap is retrieved by other methods, our CMMLoc achieves more accurate localization performance. In case (c), although some of the top-3 submaps retrieved by our coarse submap retrieval are negative, CMMLoc effectively localizes the text queries within a 10m range after applying the fine localization network. Moreover, we present a failure case in (d), where all retrieved submaps are negative. In this case, the retrieved submap contains objects with semantic labels and categories that closely resemble those in the ground truth. Consequently, the submap features modeled using the Cauchy Mixture Model are insufficiently discriminative, highlighting the importance of developing more robust representations to better distinguish between submaps.
%
\end{document}